\documentclass[letterpaper, 10 pt, conference]{ieeeconf}  % Comment this line out if you need a4paper

\IEEEoverridecommandlockouts                              % This command is only needed if 
\usepackage{graphicx}
\usepackage{amsmath}
\usepackage{amssymb}
\usepackage{hyperref}
\usepackage{multirow}
\usepackage{booktabs}
\usepackage{makecell}
\usepackage{siunitx}
\usepackage{caption}
\let\labelindent\relax
\usepackage{enumitem}
\usepackage[font=small]{caption}
\usepackage{placeins}

\usepackage[table,dvipsnames]{xcolor}
\usepackage[colorinlistoftodos, disable]{todonotes}

\usepackage[final]{changes}
\definechangesauthor[color=blue]{rev}
\setaddedmarkup{\textcolor{blue}{#1}}
\setdeletedmarkup{\textcolor{red}{\sout{#1}}}
\definecolor{g2aRed}{RGB}{255,235,235}
\definecolor{rlGreen}{RGB}{235,255,235}
\definecolor{analyticalBlue}{RGB}{235,240,255}

\title{\textbf{Grasp to Act: Dexterous Grasping for Tool Use in Dynamic Settings}
\vspace{-4mm}
}

\author{
Harsh Gupta$^1$\quad Mohammad Amin Mirzaee$^1$\quad Wenzhen Yuan$^1$ \thanks{$^1$University of Illinois Urbana-Champaign}
\\
{\fontsize{10pt}{11pt}\selectfont \texttt{\href{https://grasp2act.github.io/}{grasp2act.github.io}}}
\vspace{1mm}
}

\begin{document}
\makeatletter
\let\@oldmaketitle\@maketitle
    \renewcommand{\@maketitle}{\@oldmaketitle
    % \vspace{-2mm}
    \centering
    \includegraphics[width=1.0\textwidth]{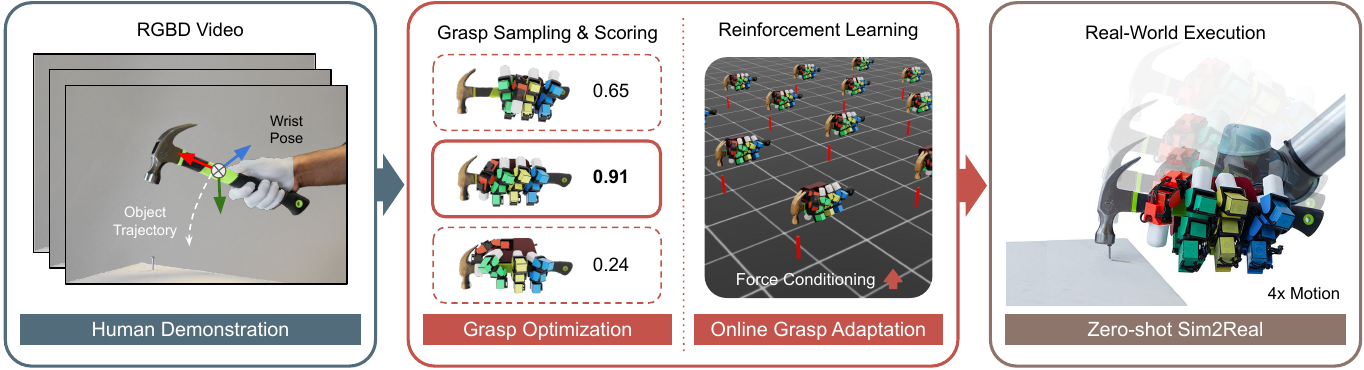}
    \captionof{figure}{\textit{Overview of the Grasp-to-Act framework.} 
    (Left) A human demonstration trajectory provides the reference of the dynamic conditions during tool using, as well as the initial grasping locations. 
    % From a single human demonstration (left), we extract the object’s 6D trajectory and the initial wrist pose using RGBD video. These are used to guide 
    Accordingly, we sample a series of grasps and score them 
    % grasp sampling and scoring 
    to identify stable grasp candidates (middle left). A reinforcement learning policy then performs online-grasp adaptation by adjusting finger joint positions under task-specific force conditioning (middle right). The resulting policy is zero-shot deployed in the real world (right).
    }
    % \vspace{-3.3mm}
    \vspace{-1mm}
    
    \label{fig:1_teaser}
    \setcounter{figure}{0}
  }
\makeatother

\maketitle

\begin{abstract}
% Dexterous robotic manipulation 
% Grasping with dexterous hands remains a core challenge for robot manipulation. 
% enabling robots to perform practical tasks in real-world environments. 
% Most existing grasping methods focus on static settings and rely primarily on object geometry, with limited consideration of the dynamic challenges that arise during manipulation. However, such dynamic conditions are common in dexterous tasks like tool use.
% simplified benchmark tasks with limited utility, struggling to achieve the stability required for tool use under dynamic conditions. 
% To address the gap in the requirement for stable grasping under dynamic conditions, we introduce \textit{Grasp-to-Act}, a hybrid framework that combines physics-based grasp optimization with reinforcement learning-based adaptive control to achieve stable grasps for real-world functional tasks. Our method synthesizes robust grasp configurations informed by human demonstrations and employs an adaptive controller that dynamically refines grasps to maintain task alignment. We validate our approach across five real-world tasks—hammering, sawing, cutting, stirring, and scooping—improving grasp stability and task success over purely analytical or learning-based approaches. Extensive experiments in both simulation and real-world settings demonstrate strong task performance, advancing the state of the art in dexterous robotic tool-use. %Results videos can be found at \href{https://sites.google.com/view/grasp2act}{sites.google.com/view/grasp2act}.
% \harsh{rewrote abstract}
Achieving robust grasping with dexterous hands remains challenging, especially when manipulation involves dynamic forces such as impacts, torques, and continuous resistance—situations common in real-world tool use. Existing methods largely optimize grasps for static geometric stability and often fail once external forces arise during manipulation.
We present \textit{Grasp-to-Act}, a hybrid system that combines physics-based grasp optimization with reinforcement-learning-based grasp adaptation to maintain stable grasps throughout functional manipulation tasks. 
Our method synthesizes robust grasp configurations informed by human demonstrations and employs an adaptive controller that residually issues joint corrections to prevent in-hand slip while tracking the object trajectory.
\textit{Grasp-to-Act} enables robust zero-shot sim-to-real transfer across five dynamic tool-use tasks—hammering, sawing, cutting, stirring, and scooping—consistently outperforming baselines. Across simulation and real-world hardware trials with a 16-DoF dexterous hand, our method reduces translational and rotational in-hand slip and achieves the highest task completion rates, demonstrating stable functional grasps under dynamic, contact-rich conditions.
\end{abstract}

\section{Introduction}

Grasping is the first step toward manipulation and remains a long-standing challenge in robotics~\cite{bohg2013data}. Most existing work focuses on identifying grasp locations based on object geometry, typically defining success as the ability to lift or hold an object stably in mid-air~\cite{wang2023dexgraspnet, lum2024get, chen2025web2grasp}. Such approaches often neglect the object's weight and dynamic interactions. In real-world manipulation, however, grasping must account for diverse object properties and task requirements. Tool use exemplifies this challenge—tasks such as hammering a nail or cutting wood with a saw involve heavy objects and asymmetric grasping configurations that generate significant torque on the hand. Moreover, these tasks introduce substantial disturbance forces and torques, whether from object dynamics (e.g., swinging a hammer) or contact interactions (e.g., impact during nail driving or resistance during sawing). These factors place much higher demands on grasp stability.

In such scenarios, dexterous hands are essential. While low degree of freedom (DoF) grippers—such as parallel-jaw or suction grippers—are widely used, their limited contact points restrict their ability to counteract large torques and maintain stable grasps under strong disturbances. Dexterous hands, in contrast, can form power grasps in which the palm provides a broad contact area to balance torque while the fingers wrap around the object to prevent slipping. Despite their potential, dexterous-hand grasping remains a difficult and relatively underexplored problem. The high dimensionality of these hands makes planning robust grasps challenging, and existing studies primarily emphasize geometric grasp location selection~\cite{miller2004graspit, liu2021synthesizing, ciocarlie2007dexterous}, with limited investigation into stability under large external loads.

To tackle these challenges, grasp planning for dexterous hands must explicitly account for grasp stability under realistic conditions, incorporating the effects of object weight, grasp-induced torque, and disturbances from both object dynamics and environmental interactions.

We propose \textit{Grasp-to-Act (G2A)}, a hybrid framework designed to ensure stable grasps during dynamic, functional tasks. \textit{G2A} introduces a stability evaluation protocol that tests candidate grasps under extreme forces and torques to identify those robust enough for contact-rich manipulation. The protocol is guided by human demonstrations—robots reproduce human motion trajectories during real-world tool-use tasks, while the resulting impedance forces and torques on the objects are modeled to assess grasp stability.

Our \textit{G2A} pipeline consists of two stages for the grasp-to-act process: an initial grasp, where we sample a set of candidate grasp configurations in a physics-based simulation platform and evaluate their stability under various disturbance forces; an online adaptation phase, when performing the human-demonstrated trajectory, where we develop an RL-based controller to counteract unexpected in-hand motion of the objects.

We validate our method across five representative functional manipulation tasks—hammering, sawing, cutting, stirring, and scooping—using a simulated and real-world robot platform equipped with a LEAP hand~\cite{shaw2023leaphand}. Our experiments show that combining grasp optimization with online-grasp adaptation significantly improves grasp stability and task success compared to baselines that rely solely on static optimization or RL.
\textit{G2A} represents a shift from grasping as a static objective to grasping as an enabling step for manipulation, moving toward robots that can perform purposeful, task-driven interaction with the physical world.
\section{Related Work}
\subsection{Dextrous Grasp Synthesis}
\label{sec:related}

Dexterous grasp synthesis aims to determine stable contact configurations for multi-fingered robotic hands given an object mesh, point cloud, or other perceptual input. 

Classical approaches~\cite{miller2004graspit, ciocarlie2007dexterous, berenson2008grasp} optimize analytical metrics such as force closure~\cite{liu2021synthesizing}, form closure~\cite{wu2022learning}, min-weight metric~\cite{li2023frogger}, or Ferrari–Canny metric~\cite{ferrari1992planning} using sampling-based or gradient-based methods. However, these methods suffer from two key limitations. First, they rely on accurate object models and often make strong contact assumptions~\cite{bohg2013data}, such as perfect rigidity, point contacts, zero or uniform friction, and a quasi-static setting (i.e. no contact breakage). \replaced{Second, classical methods often optimize for fingertip contacts resulting in precision grasps, though recent work such as Dexonomy~\cite{chen2025dexonomy} has demonstrated synthesis across broader grasp taxonomies~\cite{feix2015grasp}. However, these methods evaluate grasps under static or gravitational conditions rather than the large task-specific wrenches encountered during dynamic tool use}{Second, they typically optimize only for fingertip contact points that result in precision grasps, overlooking the broader range of grasp taxonomies~\cite{feix2015grasp}}. In practice, these methods are not applicable to heavy or irregularly shaped objects such as tools. To address these concerns, we build our grasp synthesis pipeline entirely within a physics-based simulator and employ stochastic sampling to generate a diverse set of candidate grasps across functional regions of the object.

Data-driven methods~\cite{wang2023dexgraspnet, lum2024get, Xu_2023_CVPR} for dexterous grasp synthesis typically rely on large-scale datasets~\cite{li2022gendexgrasp} of synthetic grasps paired with perceptual inputs, often generated using the grasp optimization techniques described earlier. Because these optimization frameworks frequently synthesize physically infeasible grasps, many datasets apply a post 
% \amin{what is hoc filter? I could search for ad-hoc filter. I think citation is needed} \harsh{from google: A "post hoc filter" is a generic term for any filtering or data processing that is performed after an initial analysis has been completed.}
hoc filtering step using physics-based simulators to discard unstable grasps, typically through a simple 1D or 6D gravitational test. \replaced{Recent work has explored ranking grasps using continuous stability scores~\cite{robinson2021contrastive, casas2024multigrippergrasp, lum2024get}, but these evaluation strategies remain limited to gravitational stability}{Recently, efforts have focused on retaining and ranking both successful grasps and high-quality failed examples~\cite{robinson2021contrastive}, driven by the strong performance of discriminative models for parallel-jaw grasping~\cite{Mousavian_2019_ICCV}. MultiGripperGrasp~\cite{casas2024multigrippergrasp} addresses this by assigning a continuous score based on the object's fall-off time under gravity. Get a Grip~\cite{lum2024get} simulates grasps under small perturbations and averages the outcomes to estimate the probability of success. However, both evaluation strategies are limited to testing gravitational stability}. Drawing from these methods, we design a set of rigorous wrench-space stability tests that evaluate our synthesized grasps under extreme forces and torques, enabling us to rank and select the most stable candidates for downstream tasks.

\subsection{Reinforcement Learning for Dexterous Manipulation}

Reinforcement learning (RL) has demonstrated strong potential in dexterous manipulation, achieving impressive results in tasks such as pick-and-place, in-hand reorientation, and object relocation~\cite{qi2023general, singh2024dextrah, chen2023visual}. \replaced{However, these tasks primarily involve isolated object motion and do not account for the external forces and torques that arise in contact-rich manipulation, limiting their applicability to dynamic tool-use}{However, these tasks primarily involve isolated object motion and do not account for the external forces, impacts, and torques that arise in real-world tool-use and contact-rich manipulation. As a result, existing RL policies are not designed to maintain grasp stability under dynamic interactions with the environment. Furthermore, they often rely on extensive training, carefully shaped task-specific rewards, and heavy domain randomization to achieve sim-to-real transfer—factors that limit their scalability to dynamic force-intensive tasks}.

To improve sample efficiency and encourage physically meaningful behaviors, many works incorporate human priors by initializing RL policies from human motion data or pre-grasp configurations~\cite{rajeswaran2017learning, lum2025crossing, huang2025fungrasp}. While such priors accelerate learning, they cannot guarantee grasp stabilitys once dynamic forces act on the object. Other approaches reduce the action dimensionality using eigengrasp-based controllers~\cite{ciocarlie2007dexterous, agarwal2023dexterous, lum2024dextrah}, but the resulting low-rank control spaces limit the fine-grained torque regulation required for dynamic grasp adaptation.

In contrast, our method employs RL not to learn full manipulation behaviors but to adapt grasp stability online during tool-use trajectories. By initializing the policy from analytically stable grasps and training on residual joint corrections, our approach mitigates the impact of inaccurate contact dynamics~\cite{wang2024lessons, salvato2021crossing} and improves robustness under strong external forces.
\section{Method}
\setcounter{figure}{1}
\begin{figure*}[t]
    \centering
    \includegraphics[width=\linewidth]{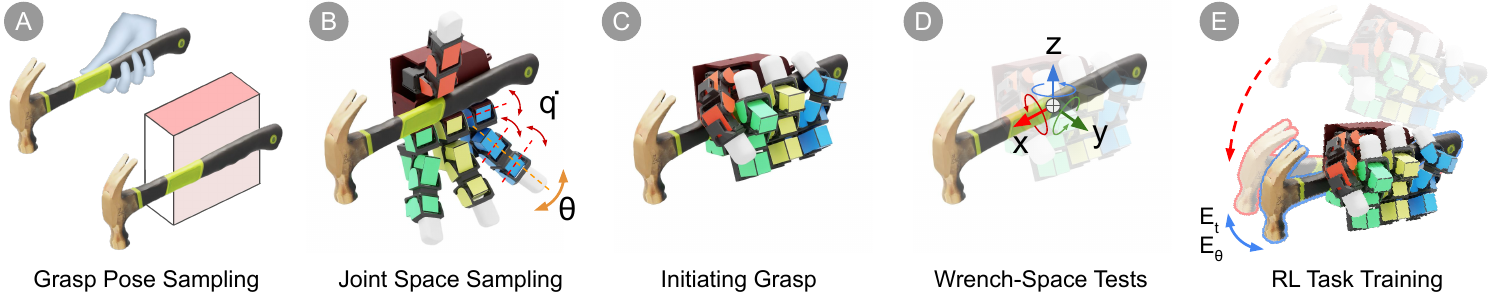}
    \caption{
    Pipeline of sampling and evaluating grasps. (A) We initialize the range of grasp locations based on human demonstration. (B-C) Candidate grasps are generated by varying joint-level parameters.
    % \textit{Grasp synthesis and evaluation pipeline.} (A) Human wrist pose from demonstration guides grasp region sampling. (B-C) Candidate grasps are generated by varying joint-level parameters. 
    (D) Each grasp is evaluated through wrench-space stability tests along six force and torque axes. (E) Top-ranked grasps are used as the initial grasp for conducting the tool-using trajectory, followed by online RL-based adaptation. 
    % for RL-based online adaptation.
    }
    \label{fig:opt_method}
    \vspace{-5mm}
\end{figure*}
% \subsection{Overview}

% To enable robots to perform functional, human-like tasks, 
To enable robots to succeed in dexterous grasping tasks, grasps must not only be stable but also functionally appropriate (e.g., holding a hammer by the handle rather than by the head). Furthermore, for successful task execution, the robot hand must maintain this stable grasp while following the task-specific trajectories. Our framework integrates 
% demonstration-guided 
sampling-based grasp synthesis with online reinforcement learning–based adaptation, as illustrated in Fig.~\ref{fig:opt_method}. 

\subsection{Human Demonstration}
We begin by recording an RGBD video $\{I^t\}_{t=1}^T$ of a human demonstrating the process of grasping the target tool and performing the task. This demonstration data is used for two purposes: initializing the search space for the task-driven grasping, and providing a reference trajectory of moving the object in grasp. 

% We begin by recording a single RGBD video $\{I^t\}_{t=1}^T$, where each frame $I^t \in \mathbb{R}^{H \times W \times 4}$ captures a human performing the task. 
Based on each frame in the video, we extract the object’s 6D pose trajectory using FoundationPose~\cite{wen2024foundationpose}, conditioned on the object's mesh, the full RGB-D video, and a segmentation mask of the object in the first frame. The initial mask is obtained using Grounded-SAM~\cite{ren2024grounded}, prompted with the object name and the initial RGB image without occlusions. This process produces a per-frame object pose sequence $\{\mathbf{T}_{\text{obj}}^t\}_{t=1}^T$, where $\mathbf{T}_{\text{obj}}^t \in \text{SE}(3)$. The object trajectory is used to define a task-specific reward for trajectory tracking during policy learning.

To localize the human's grasp, we annotate the timestep $\tau$ at which the object begins task motion. At this frame, we estimate the 6D wrist pose \(\mathbf{T}_{\text{wrist}}^\tau \in \text{SE}(3)\) in the camera frame using the pipeline introduced by Lum et al~\cite{lum2025crossing}. Specifically, we use an off-the-shelf hand pose estimator, HaMeR~\cite{pavlakos2024reconstructing}, to obtain a detailed representation of the hand~\cite{romero2022embodied}, then refine the result using ICP alignment between the predicted hand mesh and the observed depth image $I^\tau$. The resulting wrist pose \(\mathbf{T}_{\text{wrist}}^\tau\) is used to guide downstream grasp synthesis.

\subsection{Grasp Synthesis and Evaluation}

\setcounter{figure}{1}
To generate an initial grasp, we sample a set of feasible candidate grasp configurations and evaluate their stability. Based on a stability score, we select the optimal grasp configuration to execute the trajectory. This process is performed in simulation (Isaac Lab~\cite{mittal2023orbit}).

\subsubsection{Grasp Generation}

% We overhaul the grasp synthesis framework by adopting a physics-based simulation built on Isaac Lab~\cite{mittal2023orbit}. 
We define a grasp configuration as \( G = \{q_o, \mathbf{T}_{\text{grasp}}\} \), consists of the \replaced{commanded finger joint targets $q_o \in \mathbb{R}^{n}$ (PD setpoints)}{robot's finger joint angles \(q_o \in \mathbb{R}^{n}\)} and the wrist pose \(\mathbf{T}_{\text{grasp}} \in SE(3)\) in the object frame. During the grasp formation process, the object is suspended in mid-air.
The grasp generation procedure involves two primary steps:

\paragraph{Functional Wrist Region Sampling}

We define the grasp region based on the human wrist pose \(\mathbf{T}_{\text{wrist}}^{\tau}\). \added{Rather than directly retargeting the human finger configuration---which is unreliable given the morphological differences between human and robot hands (e.g., fewer fingers, different segment proportions)---we sample diverse candidates around the demonstrated wrist pose, allowing the robot to discover grasps suited to its own morphology.} Candidate wrist poses \(\mathbf{T}_{\text{grasp}}\) are uniformly sampled within a predefined region centered on \(\mathbf{T}_{\text{wrist}}^{\tau}\). Formally, \(\mathbf{T}_{\text{grasp}} = \mathbf{T}_{\text{wrist}}^{\tau} \cdot \mathbf{T}_{\text{perturb}},\)
where \(\mathbf{T}_{\text{perturb}}\) includes translations \(\Delta \{x,y,z\}\) and rotations \(\Delta \{\theta_x, \theta_y,\theta_z\}\), each sampled uniformly within specified limits, represented by the box in Fig.~\ref{fig:opt_method}A. 

\paragraph{Finger Configuration Sampling}

To synthesize diverse grasps that can robustly conform to varied object geometries, we sample several parameters controlling the initial finger configurations (shown in  Fig.~\ref{fig:opt_method}B) and dynamics of finger closing:

\begin{itemize}
    \item \textit{Inter-finger Angles} (\(\theta_f\)): Angles between adjacent fingers \(f\) are uniformly sampled to directly control how spread apart these fingers are at grasp initiation. The thumb angle \(\theta_t\) is sampled separately based on its initial configuration. Varying these angles enables the grasp to accommodate objects of different widths and shapes. 

    \item \textit{Joint Group Closing Rates} (\(\dot{q}_g\)): \replaced{Each finger's joint groups (proximal, intermediate, distal) are assigned independent closing speeds, enabling diverse grasp shapes from claw-like to enveloping power grasps.}{Each finger is divided into hierarchical joint groups (proximal, intermediate, distal segments), and each group \(g\) is assigned its own randomized closing speed. By independently varying these closing rates, the fingers can form distinctly different grasp shapes. For example, faster distal joints with slower proximal joints lead to claw-like grasps, while synchronized rates among all joints produce enveloping, power-like grasps.}

    \item \textit{Joint Torque Threshold} ($\tau_q^*$)
    % \(q_{\tau}\))
    : Finger joints stop closing when their exerted joint torque exceeds a uniformly sampled threshold $\tau_q^*$\added{; the commanded target $q_o$ is recorded at this point}. Randomizing this threshold leads to grasps with different levels of tightness, which helps the hand adapt to objects with varying shapes and surface features. This variation makes it easier to generate grasps that can handle surface irregularities or misalignments during grasp formation.
\end{itemize}

After the grasping is complete, the object is unfrozen, allowing it to settle naturally under contact and gravity (shown in Fig.~\ref{fig:opt_method}C). 
% The resulting grasp configuration \(G\) is recorded for stability evaluation.

\subsubsection{Grasp Stability Scoring}

Each generated grasp configuration \(G\) is evaluated for stability through a set of disturbance wrench tests. Here, the wrench-space comprises all Cartesian force and torque axes in both positive and negative directions ($\pm x,\pm y,\pm z$), totaling 12 independent test dimensions, Fig.~\ref{fig:opt_method}D. In each wrench-space test dimension \(i\), the magnitude of the external wrench is increased linearly from zero up to a predefined maximum force \(F_{\text{max}}\) or torque \(\tau_{\text{max}}\) for a fixed duration \(t_{\text{max}}\). In-hand slip is defined as object deviation beyond position or orientation thresholds (\(\delta_p\), \(\delta_\theta\)). When a slip occurs, the test is terminated, and we record the elapsed stable time \(t_{\text{stable}}^{(i)}\) for that direction. The grasp stability score \(S_G\) is then computed by averaging the stable time fractions across all wrench-space dimensions:
\(
S_G = \frac{1}{12}\sum_{i=1}^{12}\frac{t_{\text{stable}}^{(i)}}{t_{\text{max}}}.
\)

Before each test, the grasp configuration \(G\) is reinstated to ensure consistent initial conditions. The grasps are ranked according to their stability score \(S_G\), and the highest-ranked grasps $G^*$ are selected and stored for online grasp adaptation. The grasp synthesis and evaluation pipeline is parallelized and can identify hundreds of usable grasps and multiple robust grasps within 2 minutes.

\subsection{Online Grasp Adaptation}
\label{sec:post_grasp}
\replaced{During task execution, the robot wrist replays the demonstrated object trajectory open-loop while a reinforcement learning (RL) policy $\pi_\theta$ issues residual finger joint corrections to maintain grasp stability under dynamic forces.}{After performing the initial grasp, we implement a reinforcement learning (RL) policy $\pi_\theta$ focused on fine-tuning finger joint movements to maintain grasp stability throughout dynamic task execution.} We formulate this as a grasp refinement problem from optimized grasps $G^*$, where the joint angles are initialized to $q_o$, and the grasp pose $\mathbf{T}_{\text{grasp}}$ is kept fixed in the object's frame. Formally, $\mathbf{T}_{\text{hand}}^t = \mathbf{T}_{\text{obj}}^t \cdot \mathbf{T}_{\text{grasp}}$, giving us the location of the robot wrist at timestep $t$.

\subsubsection{Simulation Environment}

For efficient training and scalability across diverse tasks, we develop a simulation environment that abstracts complex physical interactions into simplified force-based models. Tasks that involve interactions with fluids or deformable materials, such as stirring pancake batter or cutting cucumbers, are computationally expensive to simulate directly. To overcome this limitation, we approximate these interactions using \added{online} forces \deleted{strategically} applied at relevant object contact points\added{, computed from the current simulation state at each timestep}. We categorize these forces into two types:
\begin{itemize}
\item \textit{Resistive Forces} ($F_{\text{res}}$): These forces act in the opposite direction of the intended tool motion, simulating frictional and resistive interactions. Formally, \(F_{\text{res}} = k_{\text{res}} \cdot v_{\text{eff}},\) where \( k_{\text{res}} \) is the resistive coefficient and \( v_{\text{eff}} \) is the local velocity vector. For example, a resistive force is applied to the end effector of the spoon while stirring pancake batter. 
\item \textit{Application Forces} ($F_{\text{app}}$): These are modeled as directional forces applied onto the object (e.g., payload, normal). Formally, \( F_{\text{app}} = k_{\text{app}} \cdot \hat{d},\) where \( k_{\text{app}} \) is the application coefficient and \( \hat{d} \) is the task-specific unit direction. For instance, a continuous downward force simulates the weight of the scooped item, turning off upon dumping.
\end{itemize}

\replaced{We calibrate force coefficients from real-world measurements: resistive forces are measured using a portable force gauge by moving the tool through the material at task speeds, and payload forces are set from the measured mass of carried items. For hammering, we instead model the nail as immovable and let contact dynamics generate task disturbances. Object masses are measured with a digital scale; inertias are derived from the mesh assuming uniform density. Friction coefficients are set based on material pairs. During RL training, we domain-randomize mass, friction, and force coefficients ($\pm 30\%$) to improve robustness to real-world variation.}{Real-world payloads and contact forces are measured during task execution. We match our measured values to the simulation by tuning our force coefficients ($k_{\text{res}}, k_{\text{app}}$) on a per-task basis. Further, we apply domain randomization to our coefficients to improve performance under real-world conditions.}

\subsubsection{Reinforcement Learning Formulation}

We define a reinforcement learning policy $\pi_\theta$ parameterized by $\theta$ as follows:
\(\pi_\theta: s_t \mapsto \Delta q_{t},\) where the observation state $s_t$ is given by the concatenation of the robot's proprioceptive state $s_{\text{robot}}$, the \replaced{target object pose $s_{\text{goal}} \in \mathbb{R}^7$ from the demonstrated trajectory}{goal object state $s_{\text{goal}}$}, and a normalized task timestep $t/T$. Here, the robot's proprioceptive state includes joint positions $q_t \in \mathbb{R}^n$, target joint positions $q_t^{\text{target}} \in \mathbb{R}^n$, and joint torques $\tau_t \in \mathbb{R}^n$. \added{Instead of directly observing the object pose, the policy infers the object's state indirectly through hand joint positions, targets, and torques; when the object shifts in hand, these proprioceptive signals change, providing sufficient feedback for grasp adaptation.} The output residual joint adjustments $\Delta q_t \in \mathbb{R}^n$ represent incremental corrections to the current joint configuration. Formally, $q_{t+1} = q_t + \Delta q_t$, where $q_{t+1}$ are the executed joint positions. 

The reward function combines separate terms for position alignment, orientation alignment, and penalties for large deviations. We define the position reward as using a mean square error:
\[
r_{p}^t = \max\left(1 - \alpha_p \cdot ||p_{\text{obj}}^t - p_{\text{goal}}^t||_2, 0\right),
\]
where $p_{\text{obj}}^t, p_{\text{goal}}^t \in \mathbb{R}^3$ are the current and target object centroid positions, and $\alpha_p$ is a scaling coefficient. We define the orientation reward using a quaternion dot product:
\[
r_{q}^t = \max\left(1 - \alpha_q \cdot (1 - \left| q_{\text{obj}}^t \cdot q_{\text{goal}}^t \right|),\, 0\right)
\]
where $q_{\text{obj}}^t, q_{\text{goal}}^t \in \mathbb{S}^3$ are the current and target object quaternions, and $\alpha_q$ is a scaling coefficient. A large negative penalty $r_{\text{pen}}^t$ is applied if the position or orientation exceeds predefined thresholds $\delta_p$ for position or $\delta_\theta$ for orientation. Episodes are reset after the penalty is applied. The final reward at timestep $t$ is given by:
\(
r^t = r_p^t + r_q^t + r_{\text{pen}}^t.
\)

We implement our policy $\pi_\theta$ using proximal policy optimization (PPO) \cite{schulman2017proximal}. Our network consists of an input layer that receives the state $s_t$, followed by a 512-unit LSTM \cite{hochreiter1997long} layer, an MLP with hidden sizes 512 and 256, and an output layer that predicts the residual joint adjustments $\Delta q_t$.

Given optimized grasps $G^*$, our simple reward formulation is sufficient to guide the policy to converge in $\sim$15 minutes on a single NVIDIA RTX 4070 Ti Super.
% Additional rewards did not meaningfully increase task performance or training speed. 
 
\section{Experiments}

Our experiments seek to validate whether \textit{Grasp-to-Act} enables dexterous robot hands to maintain stable, functional grasps during dynamic and contact-rich manipulation tasks. The proposed tasks pose diverse challenges, requiring grasps to maintain low positional and angular errors, while facing repetitive impacts or continuous external forces.

\subsection{Hardware Setup}

\setcounter{figure}{2}
\begin{figure}
    \centering
    \includegraphics[width=1.0\linewidth]{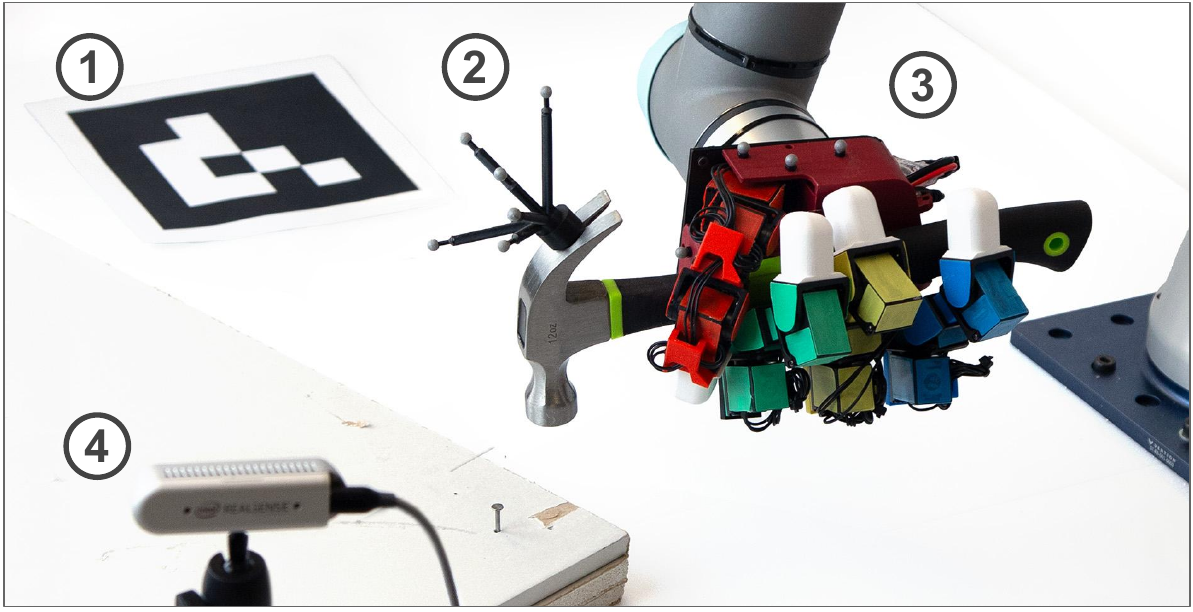}
    \caption{\textit{Real-world experiment setup.} 
    The workspace includes an (1) AprilTag, (2) tool tracking markers, (3) hand tracking markers, and an (4) RGBD camera.}
    \label{fig:placeholder}
    \vspace{-5mm}
\end{figure}

We deploy our pipeline using a 16-DoF LEAP hand~\cite{shaw2023leaphand} mounted on a 6-DoF UR5 robotic arm, and a stationary Intel RealSense D435 RGB-D camera. The execution procedure in the real world involves the following steps:

We identify the object's initial pose relative to the camera frame $\mathbf{T}_{\text{cam} \leftarrow \text{obj}} \in \text{SE}(3)$ using FoundationPose~\cite{wen2024foundationpose}. We use an AprilTag~\cite{olson2011apriltag} fixed in the workspace to determine the camera's pose relative to the global coordinate frame $\mathbf{T}_{\text{world} \leftarrow \text{cam}} \in \text{SE}(3)$.  Given our optimized grasp's wrist pose relative to the object's frame $\mathbf{T}_{\text{grasp}}$, we can compute the grasp location in the world frame $\mathbf{T}_{\text{world} \leftarrow \text{grasp}}$ by chaining the transformations:
$\mathbf{T}_{\text{world} \leftarrow \text{grasp}} = \mathbf{T}_{\text{world} \leftarrow \text{cam}} \cdot \mathbf{T}_{\text{cam} \leftarrow \text{obj}} \cdot \mathbf{T}_{\text{grasp}}$

The UR5 arm moves the robot wrist to \(\mathbf{T}_{\text{world} \leftarrow \text{grasp}}\) via inverse kinematics (IK), after which the LEAP hand \replaced{closes toward the synthesized PD targets $q_o$; the fingers stop when contact forces balance the PD effort, producing an enveloping grasp. For the sawing task, where fingers must pass through a handle loop, we manually stage the object with the grasp region accessible}{executes the optimized finger joint configuration \(q_o\). We assume that the grasping region is unobstructed at execution time}.

Following grasp execution, the robot moves the grasped object to align with the initial position of the demonstrated trajectory. Specifically, at each timestep $t$, the target wrist pose \(\mathbf{T}_{\text{hand}}^t\) is calculated from the human-demonstrated object trajectory $\{\mathbf{T}_{\text{obj}}^t\}_{t=1}^T$ as described in Sec.~\ref{sec:post_grasp}. IK is used to move the robot wrist to each sequential pose \(\mathbf{T}_{\text{hand}}^t\). Concurrently, the RL policy \(\pi_\theta\) issues joint adjustments \(\Delta q_t\) at each timestep. Both the wrist trajectory and finger joint angles are updated synchronously at a control frequency of 30 Hz.

We use an OptiTrack motion-capture system with a rigid four-marker mount attached to each object to accurately record its 6D pose during real-world trials (see Fig.~\ref{fig:placeholder}).

\subsection{Tasks and Objects}
\label{sec:tasks}
\begin{figure}[t]
    \centering
    \includegraphics[width=1.0\linewidth]{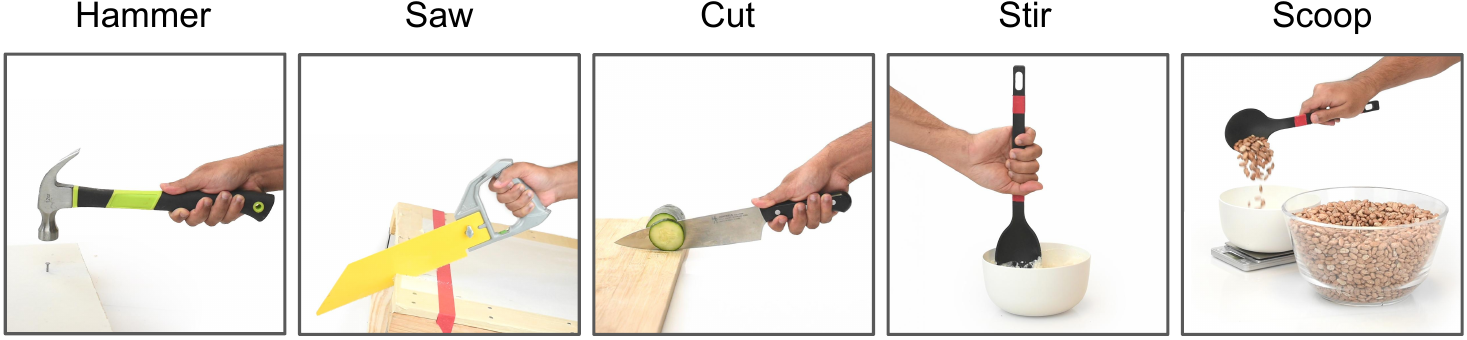}
    \caption{
    % \textit{Task Visualization.} 
    Human demonstrations of the five functional tasks we test with in this paper: hammering, sawing, cutting, stirring, and scooping.}
    \label{fig:tasks}
    \vspace{-5mm}
\end{figure}

We evaluate our approach on five manipulation tasks that require different stable grasps and dynamic control (see Fig.~\ref{fig:tasks}). These tasks involve different object geometries, force applications, and motion patterns. For each task, we also define a task completion metric $\mathcal{T}$ to quantitatively assess performance in the real world:
\begin{enumerate}[label=\textit{(\alph*)}]
    \item \textit{Hammer:} Drive a nail 2~cm above a drywall stack using a steel claw hammer (30~cm length, rubber handle, 520~g) with 4 strikes. Performance is the fraction of the nail embedded.  
    \item \textit{Saw:} Cut an 8~cm groove through drywall using a handheld saw (26~cm serrated blade, metal handle, 365~g) with 12 forward–return motions. Performance is the sawed length relative to the target.  
    \item \textit{Cut:} Slice a 5~cm cucumber segment using a stainless-steel kitchen knife (8-inch blade, plastic handle, 155~g) with 16 forward–return motions. Performance is the cut depth normalized by the cucumber diameter.  
    \item \textit{Stir:} Stir pancake batter with a plastic ladle (32~cm length, plastic handle, 60~g) for 26 clockwise rotations. Performance is the fraction of total stirring time the tool is held securely.  
    \item \textit{Scoop:} Transfer 35~g of pinto beans using the same ladle in one motion. Performance is the transferred mass ratio to the target weight.  
\end{enumerate}

\subsection{Baselines}

\setlength{\tabcolsep}{4pt}
\begin{table*}[h]
    \centering
    \label{tab:simulation_results}
    \begin{tabular}{l c c c c c c c c c c c c c c c}
        \toprule
        & \multicolumn{3}{c}{Hammer}
        & \multicolumn{3}{c}{Saw}
        & \multicolumn{3}{c}{Cut}
        & \multicolumn{3}{c}{Stir} 
        & \multicolumn{3}{c}{Scoop} \\
        \cmidrule(lr){2-4} \cmidrule(lr){5-7} \cmidrule(lr){8-10} \cmidrule(lr){11-13} \cmidrule(lr){14-16}
        Method 
        & $\mathcal{S}\uparrow$ & $E_{t}\downarrow$ & $E_{\theta}\downarrow$
        & $\mathcal{S}\uparrow$ & $E_{t}\downarrow$ & $E_{\theta}\downarrow$
        & $\mathcal{S}\uparrow$ & $E_{t}\downarrow$ & $E_{\theta}\downarrow$
        & $\mathcal{S}\uparrow$ & $E_{t}\downarrow$ & $E_{\theta}\downarrow$
        & $\mathcal{S}\uparrow$ & $E_{t}\downarrow$ & $E_{\theta}\downarrow$ \\
        \midrule

        Analytical~\cite{liu2021synthesizing, wang2023dexgraspnet, lum2024get}
        & $0.0$ & -- & --
        & $0.0$ & -- & --
        & $0.0$ & -- & --
        & $0.0$ & -- & --
        & $0.0$ & -- & -- \\

        RL base
        & $100.0$ & $0.93$ & $4.07$ 
        & $54.7$ & $6.63$ & $13.53$ 
        & $81.2$ & $0.85$ & $6.25$
        & $98.5$ & $0.51$ & $5.34$ 
        & $100.0$ & $0.52$ & $2.57$ \\

        \quad + contact reward~\cite{xu2025dexsingrasp}
        & $100.0$ & $0.86$ & $3.99$ 
        & $49.3$ & $6.92$ & $14.16$ 
        & $74.6$ & $0.94$ & $7.43$
        & $100.0$ & $0.47$ & $4.54$ 
        & $100.0$ & $0.46$ & $1.94$ \\

        \quad + pre-grasp pose~\cite{lum2025crossing}
        & $100.0$ & $0.88$ & $9.48$ 
        & $100.0$ & $2.24$ & $6.74$
        & $20.4$ & $2.20$ & $17.04$ 
        & $100.0$ & $0.45$ & $3.47$ 
        & $96.9$ & $1.23$ & $6.32$  \\

        \quad + eigengrasp~\cite{agarwal2023dexterous, lum2024dextrah, lum2025crossing}
        & $67.2$ & $1.70$ & $14.36$ 
        & $100.0$ & $1.95$ & $5.84$
        & $0.0$ & -- & --
        & $0.0$ & -- & --
        & $0.0$ & -- & -- \\

        \midrule

        G2A w/o Adaptation
        & $100.0$ & $0.82$ & $7.33$
        & $100.0$ & $1.29$ & $3.90$
        & $100.0$ & $0.54$ & $3.53$
        & $100.0$ & $0.92$ & $4.73$ 
        & $100.0$ & $0.69$ & $3.69$ \\

        G2A (ours)
        & $100.0$ & $\mathbf{0.69}$ & $\mathbf{1.35}$
        & $100.0$ & $\mathbf{1.06}$ & $\mathbf{3.13}$
        & $100.0$ & $\mathbf{0.48}$ & $\mathbf{2.23}$
        & $100.0$ & $\mathbf{0.24}$ & $\mathbf{2.08}$
        & $100.0$ & $\mathbf{0.17}$ & $\mathbf{1.25}$ \\

        \bottomrule
    \end{tabular}
    \caption{\textit{Simulation results across five tasks.} We report the grasp success $\mathcal{S}$ (\%), in-hand translational slip distance $E_t$ (cm) and in-hand slip rotation distance $E_\theta$ (\si{\degree}). All error metrics ($E_t, E_\theta$) are averaged over successful episodes.}
    \label{tab:simulation_results}
    \vspace{-5mm}
\end{table*}

We compare our method against several baselines representative of analytical optimization and RL-based approaches commonly used in dexterous manipulation tasks:

(1) \textit{Analytical~\cite{liu2021synthesizing,wang2023dexgraspnet,lum2024get}}: Grasps are synthesized using a geometric differentiable optimization method. To make this baseline suitable for our functional tasks, we initialize the optimization from human-demonstrated wrist poses near functional regions of the tool, and then rank generated grasps using \replaced{the same wrench-space stability evaluation (Sec.~III-B.2) as our method}{our evaluation protocol}.

(2-5) \textit{RL-based methods}: These baselines are given a 5~second initialization phase, where the object remains fixed in space, allowing the robot hand to change its initial wrist and finger configurations to make a grasp. The initial wrist pose is placed at the human-demonstrated wrist location. To encourage wrist alignment during this initialization phase, we incorporate a proximity reward $r_w$ that penalizes the distance between the wrist and the object. Formally written as: $r_w = \text{max}\left( 1- \alpha_w \| p_{\text{wrist}} - p_{\text{obj}} \|_2 , 0\right).$ \textit{(2) RL base:} This policy initialized from a default joint-angle configuration with no modifications. \textit{(3) RL w/ Contact Rewards~\cite{xu2025dexsingrasp}:} This policy integrates explicit finger-object proximity rewards to encourage grasping with all fingers. \textit{(4) RL w/ Pre-Grasp Pose~\cite{lum2025crossing}:} The policy initialized from a retargeted human-demonstrated finger joint configuration\replaced{---in contrast to G2A, which initializes from our synthesized grasps. This comparison isolates the contribution of our grasp synthesis stage}{, enabling the policy to refine grasp poses that closely match human grasps}. \textit{(5) RL w/ Eigengrasp~\cite{agarwal2023dexterous, lum2024dextrah, lum2025crossing}:} The policy actions are constrained within a low-dimensional (5-dimensional) eigengrasp space derived from principal component analysis (PCA) of human grasps. This reduces action-space dimensionality and guides the exploration toward human-like grasps. We directly use Lum et al's controller \cite{lum2025crossing}.

\textit{(6) G2A w/o Adaptation:} This baseline refers to the grasps synthesized by our grasp optimization pipeline without the RL-based online adaptation. \textit{(7) G2A:} This is our full method.

% In simulation, we compare all the baseline policies, while comparing three baselines in the real-wold: \textit{RL base}, \textit{G2A w/o Adaptation}, and \textit{G2A}.

\subsection{Evaluation Metrics}
We evaluate performance using three metrics shared across simulation and real-world experiments: in-hand translational slip distance $E_t$, in-hand rotational slip distance $E_\theta$, and either grasp success $\mathcal{S}$ (used in simulation) or task completion $\mathcal{T}$ (used in real-world experiments).
\subsubsection{In-hand translational slip distance $E_t$ (cm)} For episodes without object dropping, we measure the average positional deviation of the object centroid from the target trajectory when no slip occurs:
\(
E_t = \frac{1}{T} \sum_{t=1}^{T} \left\| p_{\text{obj}}^t - p_{\text{goal}}^t \right\|_2
\)
\subsubsection{In-hand slip rotation distance $E_\theta$ (\si{\degree})} Similarly, for successful episodes, we calculate the average angular deviation between the object's orientation and the target orientation when no slip occurs:
\(
E_\theta = \frac{1}{T} \sum_{t=1}^{T} 2 \cdot \arccos\left( \left| q_{\text{obj}}^t \cdot q_{\text{goal}}^t \right| \right)
\)
\subsubsection{Task performance} As the tasks are not explicitly modeled in simulation, we instead report a grasp success $\mathcal{S}$ (\%) indicating whether the object remains held throughout the episode without dropping.
In real-world experiments, we measure task completion $\mathcal{T}$ defined in Section~\ref{sec:tasks}, which directly reflects functional task success (e.g., nail inserted, material cut).
\section{Results}

\subsection{Experiments in Simulation}

We evaluate our method and all baseline methods across all tasks in simulation over 3 random seeds, reporting average performance in  Table~\ref{tab:simulation_results}. For the optimization-based baselines, we select the top three highest-scoring grasp candidates. The RL-based policies are trained for 100K environment steps per task.

\begin{table}[t]
    \vspace{2mm}
    \centering
    \setlength{\fboxsep}{1pt}  % tighten the image box
    \renewcommand{\arraystretch}{1.1}
    \setlength{\tabcolsep}{5pt}
    \begin{tabular}{lccc c}
        \toprule
        Method & $\mathcal{S}$ (\%) $\uparrow$ & $E_t$ (cm) $\downarrow$ & $E_\theta$ (\si{\degree}) $\downarrow$ & Hammer \\
        \midrule
        RL base & 100.0 & 0.93 & 4.07 & \multirow{3}{*}{\fbox{\includegraphics[width=0.103\linewidth]{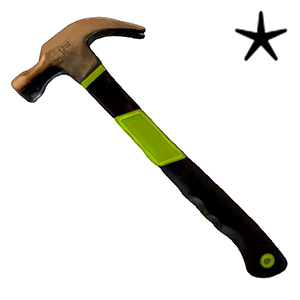}}} \\
        G2A w/o Adaptation & 100.0 & 0.82 & 7.33 & \\
        G2A (ours) & 100.0 & \textbf{0.69} & \textbf{1.35} & \\
        \midrule
        RL base & 46.6 & 2.04 & 17.12 & \multirow{3}{*}{\fbox{\includegraphics[width=0.103\linewidth]{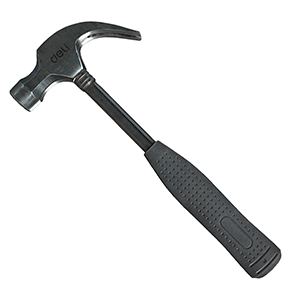}}} \\
        G2A w/o Adaptation & 92.3 & 1.02 & 11.43 & \\
        G2A (ours) & 100.0 & \textbf{0.94} & \textbf{2.69} & \\
        \midrule
        RL base & 83.7 & 1.91 & 5.98 & \multirow{3}{*}{\fbox{\includegraphics[width=0.103\linewidth]{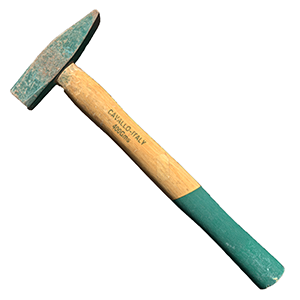}}} \\
        G2A w/o Adaptation & 100.0 & 0.91 & 9.75 & \\
        G2A (ours) & 100.0 & \textbf{0.78} & \textbf{1.57} & \\
        \bottomrule
    \end{tabular}
    \caption{\textit{Generalization across hammer types.}
    $\star$ indicates that the method was trained for that tool instance only. We report the grasp success $\mathcal{S}$ (\%), in-hand translational slip distance $E_t$ (cm) and in-hand slip rotation distance $E_\theta$ (\si{\degree}).}
    \label{tab:hammer_generalization}
    \vspace{-5mm}
\end{table}

\subsubsection{Performance Across Tasks} We first evaluate all baseline and proposed methods across the five tasks (Sec.~\ref{sec:tasks}) in simulation. 

\textit{Analytical} baseline method fails to achieve stable grasps across all tasks. \replaced{Despite using the same wrench-space ranking, the synthesized grasps are precision grasps with fingertip contacts that lack the enveloping contact needed to resist large task wrenches}{The synthesized grasps are mostly precision grasps, which suffice for picking up lighter objects such as the knife or ladle, but fail when subjected to any task-specific forces}. 

\textit{RL base} and \textit{RL w/ Contact Reward} perform similarly in simulation. Both methods often converge to suboptimal local minima, characterized by limited finger utilization and visually unstable grasp configurations that nonetheless work under simulation dynamics. Further, the lack of a human grasp prior is evident in the sawing task, where the policies grip around the handle rather than inserting fingers into it, making it difficult to maintain a secure hold. \replaced{\textit{RL w/ Pre-Grasp Pose} performs well when the human strategy is compatible with the robot embodiment (e.g., fingers within the saw handle), but does not generalize---in the knife task, the human grasp relies on the thenar eminence, which the LEAP hand lacks, leading to poor initialization.}{\textit{RL w/ Pre-Grasp Pose} baseline provides strong performance in tasks where the human strategy is compatible with the robot embodiment. For instance, initializing with fingers already within the saw handle helps maintain stability during execution. However, this benefit does not generalize across all tasks. In the knife task, the human grasp relies on the thenar eminence (the lump at the base of the thumb), which the LEAP hand lacks. As a result, direct retargeting can lead to poor initialization.} \replaced{\textit{RL w/ Eigengrasp} converges rapidly due to its lower-dimensional action space and yields human-like grasps. However, this dimensionality reduction limits dexterity for objects with smaller grasp regions, preventing fine-grained finger adjustments needed to establish sufficient contact.}{\textit{RL w/ Eigengrasp} is particularly interesting. The approach yields human-like grasping strategies, as observed in prior work. The initial training converges rapidly compared to RL-only methods, reflecting the lower-dimensional, structured action space. However, this same action space reduction limits dexterity for objects with smaller or more intricate grasp regions. As the handle size decreases, the inability to execute fine-grained finger adjustments prevents the policy from establishing sufficient contact on objects.}

\textit{G2A w/o Adaptation} baseline produces grasps that remain stable throughout the entire trajectory and do not drop the object during execution. However, under repeated application of task-specific forces, particularly in hammering, cumulative pose errors accrue over time. \added{For example, in hammering G2A w/o Adaptation achieves $E_\theta = 7.33\si{\degree}$ compared to G2A's $1.35\si{\degree}$, demonstrating the value of online RL adaptation even when starting from an already-stable grasp.} \textit{G2A} consistently outperforms all baselines. By leveraging analytically optimized grasps as initialization, the RL policy starts with a stable grasp configuration, allowing the policy to focus on minor adjustments to prevent or fix slip or rotation under dynamic task conditions. This narrows the exploration space and substantially accelerates policy convergence.

\subsubsection{Generalization across Tool Variants}
To assess robustness to unseen tool geometries, we evaluate across three hammer variants that differ in mass distribution and handle shape (Table~\ref{tab:hammer_generalization}). \added{The variants range from 460--580g in mass and 2.8--3.4cm in handle diameter.}
All policies were trained and optimized only on the first (marked with $\star$) hammer instance and directly tested on the other two without retraining.

\replaced{Baselines such as \textit{RL base} converge to a small stable region for the training tool where disturbances are limited; moderate geometric changes push grasps toward instability, causing generalization failure. \textit{G2A w/o Adaptation} maintains secure grasps on most variants but accumulates partial slip on the thinner-handled hammer. Our full \textit{G2A} method samples grasps stable under large forces and torques---a broad stability region---and adapts online, achieving $100\%$ success across all variants with minimal slip ($E_t$, $E_\theta$).}{\textit{RL base} policy performs well only on the original hammer but fails to generalize---minor shifts in handle thickness or head geometry lead to loss of contact and frequent object drops. \textit{G2A w/o Adaptation} baseline maintains secure grasps initially but cannot compensate for variations in handle diameter, resulting in partial slip for the thinner-handled hammer. In contrast, our full \textit{G2A} method consistently preserves grasp stability across all hammer variants, achieving $100\%$ success with minimal translational and rotational slip errors ($E_t$, $E_\theta$). These results highlight that the online adaptation policy effectively compensates for geometric and inertial changes.}

\subsection{Real-World Experiments}

\begin{figure}[t]
    \centering
    \includegraphics[width=1.0\linewidth]{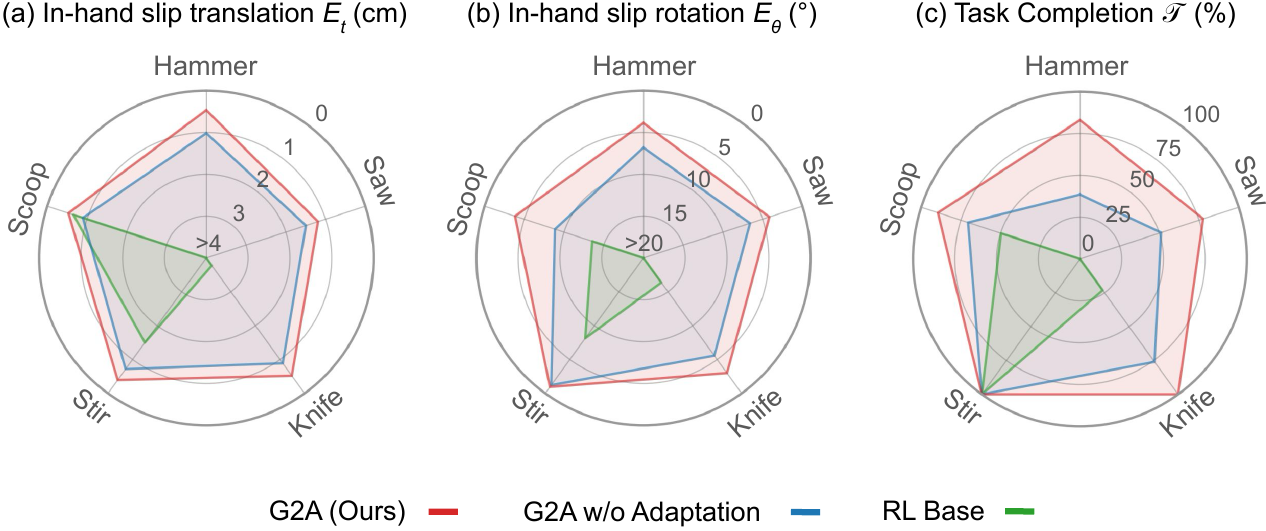}
    \caption{\textit{Real-world results across five functional tasks.} Comparison of (a) In-hand translational slip distance $E_t$ (cm), (b) In-hand slip rotation distance $E_\theta$ (\si{\degree}), and (c) task completion $\mathcal{T}$. Task-specific definitions of $\mathcal{T}$ are provided in Section~\ref{sec:tasks}.}
    \label{fig:real_world_results}
    \vspace{-5mm}
\end{figure}

\begin{figure}[t]
    \centering
    \includegraphics[width=.925\linewidth]{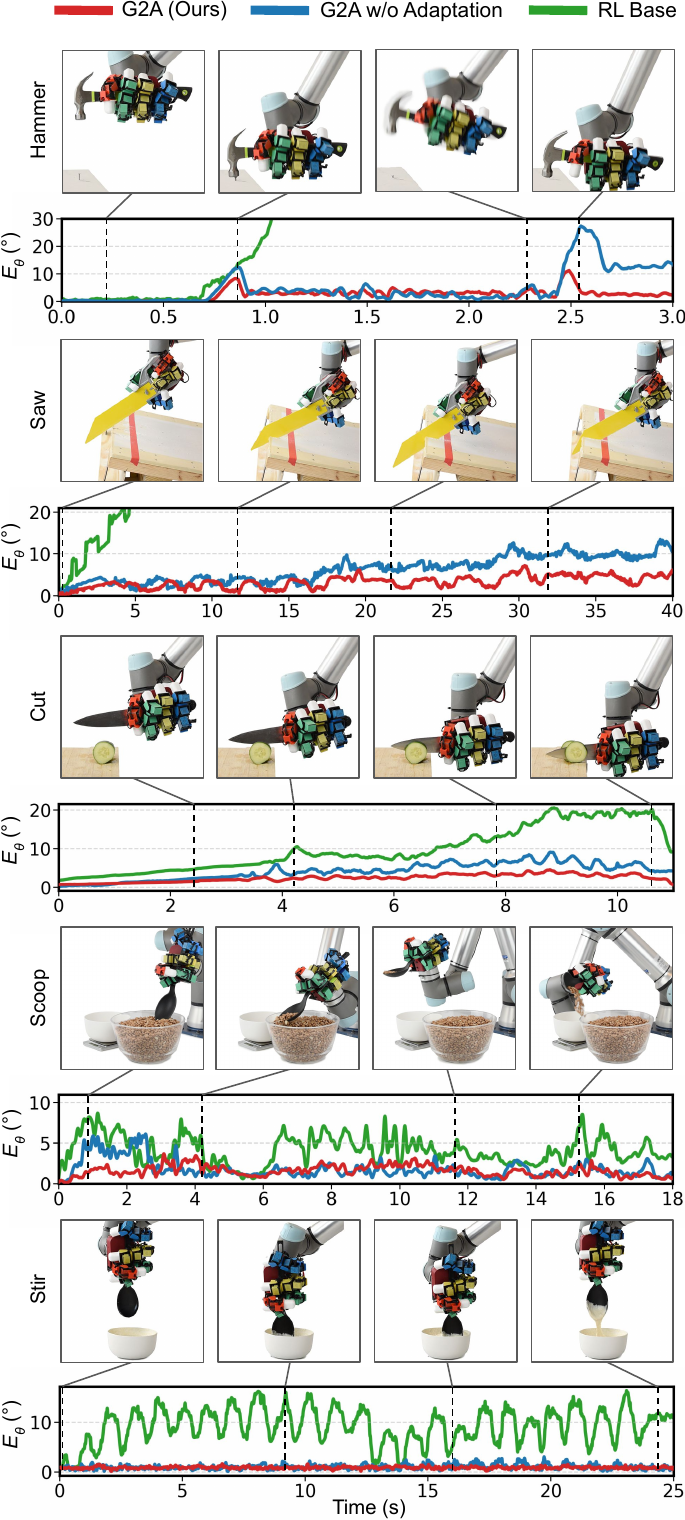}
    \caption{\textit{Real-world rollouts.} In-hand slip rotation distance $E_\theta$ for one representative rollout per baseline across the tasks. Task snapshots show the G2A execution. }
    \label{fig:3_analysis}
    \vspace{-5mm}
\end{figure}

We evaluate over 10 real-world trials per task against two baselines: \textit{G2A w/o Adaptation}, the strongest non-adaptive policy, and \textit{RL Base}, a consistent RL variant that performed reliably across most tasks. Results are shown in Fig.~\ref{fig:real_world_results}. Some examples of the experiment episodes and the rotational slip deviation is shown in Fig.~\ref{fig:3_analysis}. 

\textit{RL Base} policy performs reasonably in simulation but transfers poorly to hardware. In practice, objects such as the hammer and saw slip once large external forces are applied. Only lightweight tools can be manipulated successfully, and even then, significant orientation drift occurs.

\textit{G2A w/o Adaptation} baseline demonstrates markedly stronger sim-to-real transfer. The synthesized grasps remain stable throughout most trials, reflected in lower translation and orientation errors than \textit{RL Base}. However, as seen before, while these grasps remain secure, residual pose drift accumulates during longer or more dynamic motions, leading to moderate $E_t$ and $E_\theta$ values.
Our full \textit{G2A} method achieves the lowest translation and orientation errors across all tasks, along with the highest task completion $\mathcal{T}$. \added{The gap between G2A and G2A w/o Adaptation is most pronounced in tasks with repeated impacts (hammering) and sustained resistive forces (cutting, sawing), confirming that RL adaptation provides meaningful benefit beyond the initial grasp quality (Fig.~\ref{fig:real_world_results}).} As shown in Fig.~\ref{fig:3_analysis}, per-task orientation traces highlight the stability of G2A over time. Orientation errors remain bounded even under repeated impacts or resistive forces. The adaptive policy adjusts finger configurations in response to contact, maintaining grasp stability throughout the task.

\section{Discussion}

In this paper, we study grasp configurations and adaptive control strategies to maintain stability during real-world manipulation tasks involving significant dynamic forces. The current implementation is based on two simplifying assumptions: (1) the initial graspable region is unobstructed, and (2) the object geometries are relatively simple. 

In practice, \replaced{objects rest on tables or shelves, where contact with the surface can obstruct part of the graspable area; for sawing, we manually stage the tool with the handle loop accessible. In such cases, our framework cannot autonomously execute the desired grasp}{any objects rest on tables or shelves, where contact with the surface can obstruct part of the graspable area. In this case, our framework cannot execute
the desired grasp}. To address this, our method could be combined with adaptive grasping strategies to maneuver objects into target grasp configurations~\cite{Xu_2023_CVPR}. 

For objects with more complex geometries—though such cases are less common in dynamic, force-intensive manipulation—our sampling-based grasp synthesis could be further improved by integrating geometric grasp-planning techniques, which may enable more efficient generation and evaluation of grasp candidates. \added{Finally, our RL adaptation policies are task-specific; extending to multi-task or cross-task transfer remains an open direction.}
\section{Conclusion}
We presented \textit{Grasp-to-Act}, a framework that enables dexterous robotic hands to achieve stable and functional grasps across diverse real-world manipulation tasks. By combining grasp optimization with reinforcement learning–based adaptation, our approach maintains robust control under dynamic, contact-rich conditions. In real-world experiments, \textit{G2A} consistently outperforms baselines, achieving the highest task completion rates across a range of tasks such as hammering, sawing, cutting, stirring, and scooping. These results look to bridge the gap between simulation and practical deployment for functional dexterity.

\bibliographystyle{IEEEtran}
\bibliography{references}

\end{document}